\documentclass{article} % For LaTeX2e
\usepackage[final]{colm2026_conference}

\usepackage{microtype}
\usepackage{hyperref}
\usepackage{url}
\usepackage{booktabs}
\usepackage{enumitem}
\usepackage{hyperref}
\usepackage{tikz}
\usetikzlibrary{arrows.meta, positioning}
\usepackage{url}
\usepackage{float}
\usepackage{caption}
\usepackage{subcaption}
\usepackage{booktabs}
\usepackage{multirow}
\usepackage{graphicx}
\usepackage{comment}
\usepackage{tabularx}
\usepackage{latexsym}
\usepackage{amsmath}
\usepackage{amssymb}
\usepackage{colortbl}
\usepackage{xcolor}
\usepackage{wrapfig} % add in preamble
\usepackage{subcaption} % in preamble

\usepackage{lineno}
\usepackage{lineno}

\definecolor{darkblue}{rgb}{0, 0, 0.5}
\hypersetup{colorlinks=true, citecolor=darkblue, linkcolor=darkblue, urlcolor=darkblue}

\title{On the Non-Identifiability of Steering Vectors in Large Language Models}

\author{Sohan Venkatesh, Ashish Mahendran Kurapath  \\
Manipal Institute of Technology Bengaluru \\
\texttt{\{sohan1, ashish\}.mitblr2022@learner.manipal.edu} \\
}

\begin{document}

\ifcolmsubmission
\linenumbers
\fi

\maketitle

\begin{abstract}
Activation steering methods are widely used to control large language model (LLM) behavior and are often interpreted as revealing meaningful internal representations. This interpretation assumes that steering directions are identifiable and uniquely recoverable from input--output behavior. We show that, under white-box single-layer access, steering vectors are fundamentally non-identifiable due to large equivalence classes of behaviorally indistinguishable interventions. Empirically, we find that orthogonal perturbations achieve near-equivalent efficacy with negligible effect sizes across multiple models and traits, with pre-trained semantic classifiers confirming equivalence at the output level. We estimate null-space dimensionality via SVD of activation covariance matrices and validate that equivalence holds robustly throughout the operationally relevant steering range. Critically, we show that non-identifiability is a robust geometric property that persists across diverse prompt distributions. These findings reveal fundamental interpretability limits and highlight the need for structural constraints beyond behavioral testing to enable reliable alignment interventions.
\end{abstract}

\section{Introduction}\label{sec:intro}
 
Activation steering has emerged as a prominent technique for controlling large language model behavior \citep{turner2023steering, rimsky2024steering}, with applications spanning persona modulation \citep{chen2025persona}, representation engineering \citep{zou2023representation} and alignment research \citep{turner2024activation}. The core procedure is simple: extract a directional vector from model activations by contrasting prompts that differ along a semantic axis, then add that vector at inference time to shift model outputs in the desired direction. This simplicity has driven broad adoption across mechanistic interpretability \citep{elhage2022toy, nanda2023progress} and alignment \citep{zou2023representation}.
 
Underlying this practice is a largely unexamined assumption: that the extracted steering direction is \emph{meaningful} in a precise sense. Specifically, it is assumed that the direction is uniquely recoverable from the model's input--output behavior and that the behavioral effect of steering reflects the genuine causal role of that direction in the model's computation. Without this assumption, the interpretability value of steering is unclear. A vector that shifts outputs in the intended direction does not, by itself, reveal which internal direction is responsible for the semantic attribute in question. The claim that a steering vector \emph{represents} a concept such as honesty or formality presupposes that the recovered direction is the unique one doing the causal work.

\begin{figure*}[t]
    \centering
    \includegraphics[width=0.8\textwidth, trim=5 5 5 5, clip]{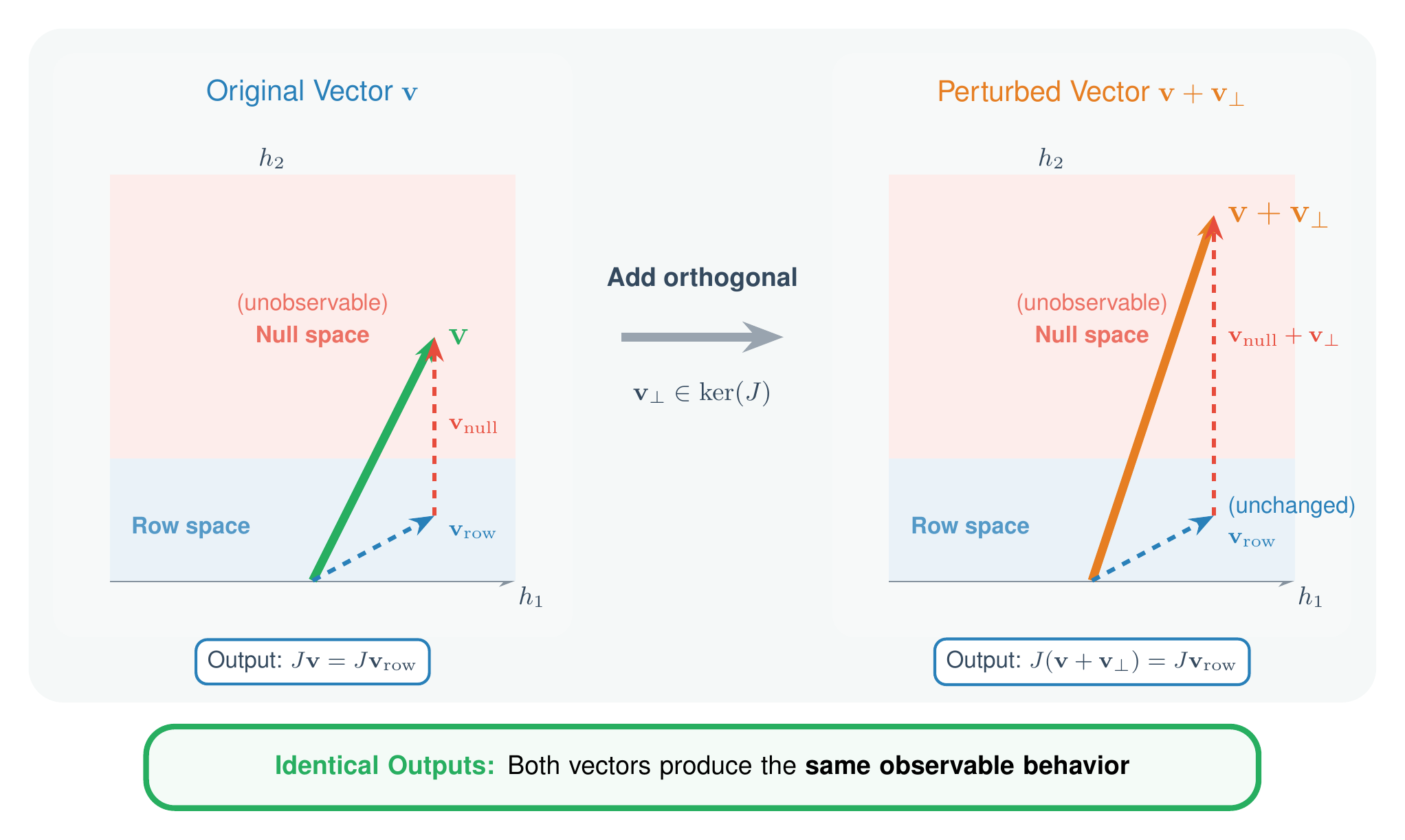}
    \caption{\textbf{Steering vector non-identifiability:} The original vector $v$ and a perturbed vector $v + v_\perp$ with $v_\perp \in \mathrm{ker}(J)$ produce identical outputs since null-space components do not affect $Jv$, the first-order change in output logits induced by steering.}
    \label{fig:introduction}
\end{figure*}

We formalize this argument and validate it empirically. The key mechanism is that the Jacobian mapping activations to output logits has a large null space in overparameterized language models \citep{oymak2019generalization}: any direction in this null space can be added to a steering vector without altering any observable output, producing infinitely many behaviorally equivalent interventions. Experiments across five semantic traits and two model families confirm that orthogonal perturbations to a steering vector are behaviorally indistinguishable from the original, that vectors extracted by entirely different methods are semantically equivalent despite being geometrically distinct and that these properties persist across diverse prompt distributions.
 
We make three contributions. First, we provide a formal identifiability analysis of steering vectors, proving non-identifiability via a constructive null-space argument under mild assumptions. Second, we demonstrate non-identifiability empirically across multiple models, traits, extraction methods, and prompt environments, using pre-trained semantic classifiers and raw logit comparisons as complementary evaluation methods. Third, we argue that the implications extend beyond steering: any intervention-based interpretability method that validates representational claims solely through behavioral output faces the same fundamental limitation and structural constraints are necessary for reliable interpretability and alignment.

\section{Related Work}\label{sec:related-work}
Our work is the first to formally analyze activation steering\footnote{We use \emph{persona vectors} and \emph{steering vectors} interchangeably to refer to directional vectors added to model activations to modulate semantic or behavioral attributes \citep{chen2025persona}.} as a latent variable identification problem, connecting two lines of research: causal representation learning and activation editing in LLMs.
 
\paragraph{Causal and latent variable identifiability.} Classical results show that latent variable models are non-identifiable without structural assumptions \citep{hyvarinen1999nonlinear, shimizu2006linear, kruskal1977three}. Recent work in causal representation learning extends these ideas to deep learning settings \citep{scholkopf2021toward, ahuja2022invariance, locatello2019challenging}, establishing conditions under which latent factors can be recovered from high-dimensional observations.
 
\paragraph{Probing and representation learning.} Work on probing classifiers \citep{pimentel2020information, ravfogel2020null, elazar2021amnesic} shows that linear directions can succeed for reasons unrelated to target information. The linear representation hypothesis \citep{park2023linear, elhage2022toy} suggests concepts correspond to directions in activation space but lacks identifiability guarantees.
 
\paragraph{Activation editing in LLMs.} Methods such as representation engineering \citep{zou2023representation}, contrastive activation addition \citep{rimsky2024steering, turner2024activation} and activation patching \citep{meng2022locating} manipulate model internals but do not address whether control directions are uniquely determined. Work on neural network symmetries \citep{entezari2021role, dinh2017sharp} motivates our null-space analysis.
 
\paragraph{Robustness and invariant representations.} Work on distribution shift and domain adaptation \citep{rabanser2019failing, lipton2018detecting} establishes that learned representations often exploit spurious correlations. Methods for learning invariant representations \citep{arjovsky2019invariant, pmlr-v139-krueger21a} aim to extract features stable across domains. Recent work on multi-environment causal representation learning \citep{ahuja2022invariance, von2021self, rojas2018invariant} proves that diverse training environments enable identification of causal factors under structural assumptions.
 
\paragraph{Mechanistic interpretability.} Work on mechanistic interpretability \citep{olsson2022context, elhage2021mathematical, nanda2023progress} studies circuits and features in transformers, often assuming the existence of interpretable directions without formalizing identifiability constraints. Our theoretical analysis clarifies when such directions are well-defined.

\section{Problem Setup}
\subsection{Formal Model}
Consider a pre-trained transformer language model $f_\theta$ with $L$ layers. For a given input prompt $x$ (tokenized as $x_1, \ldots, x_T$), let $h_\ell(x) \in \mathbb{R}^d$ denote the hidden representation at layer $\ell$ and position $T$ (typically the final token position for autoregressive generation).
 
\paragraph{Latent persona variable.} We assume there exists an underlying latent variable $z \in \mathcal{Z}$ representing a semantic attribute or "persona" (e.g., formality, political stance, truthfulness).
 
\paragraph{Steering intervention.} A steering vector $v \in \mathbb{R}^d$ is applied as:
\begin{equation}
    \tilde{h}_\ell(x) = h_\ell(x) + \alpha v
\end{equation}
where $\alpha \in \mathbb{R}$ is the steering strength. The modified representation $\tilde{h}_\ell$ is fed forward through subsequent layers to produce output logits $o(x, v, \alpha)$ over the vocabulary.
 
\paragraph{Generative model.} We posit that the true data-generating process involves:
\begin{equation}
    z \sim p(z), \quad h_\ell = g_\ell(x, z) + \epsilon
\end{equation}
where $g_\ell: \mathcal{X} \times \mathcal{Z} \to \mathbb{R}^d$ encodes how persona $z$ modulates the representation for prompt $x$ and $\epsilon \sim \mathcal{N}(0, \sigma^2 I)$ is measurement noise. The goal of steering is to approximate the effect of varying $z$ by adding $v$.
 
\subsection{Observational Regimes}
We consider two data access regimes that determine what alignment interventions can afford:
 
\begin{itemize}
\item \textbf{Regime 1: Black-box input--output.}
The researcher observes only $(x, y)$ pairs, where $y$ is generated text. There is no access to internal representations. This is the weakest regime and corresponds to behavioral evaluation.
 
\item \textbf{Regime 2: White-box single-layer access.}
The researcher can observe or manipulate activations $h_\ell(x)$ at a chosen layer $\ell$. This is the standard setting for most steering work and includes extracting vectors from contrastive prompt pairs.
\end{itemize}
 
Most existing work operates in Regime 2, often extracting a steering vector from contrastive prompt pairs $(x^+, x^-)$ designed to elicit different persona values:
\begin{equation}
    v \propto \mathbb{E}_{x^+}[h_\ell(x^+)] - \mathbb{E}_{x^-}[h_\ell(x^-)]
\end{equation}
 
Our analysis focuses primarily on Regime 2, as it represents the dominant paradigm in current steering research.
 
\subsection{Linear Approximation and Nonlinear Case}
\paragraph{Local linearization.} Near a reference distribution, we can approximate the effect of steering on output logits as:
\begin{equation}
\label{eq:first_order_expansion}
    o(x, v, \alpha) \approx o(x, 0, 0) + \alpha J_\ell(x) v
\end{equation}
where $J_\ell(x) = \left.\frac{\partial o}{\partial h_\ell}\right|_{h_\ell(x)} \in \mathbb{R}^{V \times d}$ denotes the Jacobian of the output and $V$ is the vocabulary size.
 
\paragraph{Nonlinear case.} In general, the mapping $h_\ell \mapsto o$ involves multiple nonlinear layers (attention, MLPs, layer norms). We denote this as:
\begin{equation}
    o = F_{\ell \to L}(h_\ell + \alpha v)
\end{equation}
where $F_{\ell \to L}$ is the composition of layers $\ell+1$ through $L$.
 
\subsection{Assumptions}
 
We make the following assumptions explicit.
 
\textbf{A1 (Smoothness)}
The functions $g_\ell$ and $F_{\ell \to L}$ are differentiable almost everywhere, enabling local linear approximation via Jacobians $J_\ell(x) = \frac{\partial o}{\partial h_\ell}(x)$.
 
\textbf{A2 (Identifiable prompts)}
The prompt distribution $p(x)$ has sufficient variability to probe different aspects of the latent persona $z$. Formally, the support of $p(x)$ is rich enough that different persona values induce distinguishable activation patterns $h_\ell(x)$.
 
\textbf{A3 (Non-degeneracy)}
The Jacobian $J_\ell(x)$ has rank at least $k \geq 1$ for typical $x \sim p(x)$, meaning steering can affect outputs. This excludes pathological cases where all perturbations to $h_\ell$ are ignored by subsequent layers.
 
We do \emph{not} assume: (i) statistical independence between $z$ and $x$ (confounding is allowed), (ii) linearity of $g_\ell$ in $z$ or (iii) uniqueness of the latent representation without additional structure. These assumptions are mild and generally satisfied by standard transformer architectures under typical operating conditions.

\section{Definitions and Identifiability}
\subsection{Identifiability}
 
\paragraph{Definition 1 (Parameter Identifiability).} A parameter $\theta$ in a statistical model $p(y \mid x; \theta)$ is identifiable if for any $\theta' \neq \theta$, there exists a distribution over observations $(x, y)$ such that $p(y \mid x; \theta) \neq p(y \mid x; \theta')$.
 
In our setting, the parameter is the steering vector $v \in \mathbb{R}^d$. We say $v$ is identifiable if no other vector $v' \neq v$ (up to scaling) produces the same distribution over observable outputs across all prompts and steering strengths.
 
\paragraph{Definition 2 (Observational Equivalence).} Two steering vectors $v$ and $v'$ are observationally equivalent in regime $\mathcal{R}$ if they produce identical distributions over all quantities observable in $\mathcal{R}$. For Regime 2 (white-box single-layer access):
\begin{equation}
v \sim_{\mathcal R} v' \iff F_{\ell \to L}(h_\ell(x)+\alpha v) = F_{\ell \to L}(h_\ell(x)+\alpha v') \;\forall x,\alpha .
\end{equation}
 
\subsection{Symmetries and Gauge Freedom}
\paragraph{Scaling ambiguity.} For any $c \neq 0$, the vectors $v$ and $cv$ produce outputs that differ only by a rescaling of $\alpha$. This is unavoidable; we consider $v$ and $cv$ as the same direction.
\paragraph{Null space ambiguity.} If $v_0 \in \ker(J_\ell)$ (i.e., $J_\ell v_0 = 0$), then adding $v_0$ to any steering vector does not change the linearized output. Under linear approximation, $v$ and $v + v_0$ are observationally equivalent.

\section{Main Results}\label{sec:results}
We now state our main theoretical result, proving that persona vectors are fundamentally non-identifiable under standard observational regimes.
 
\paragraph{Proposition 1. } Under Assumptions A1--A3 within the local linear approximation, in Regime 2 (white-box single-layer access) without additional structural constraints, persona vectors are not identifiable. Specifically, for any steering vector $v \in \mathbb{R}^d$, there exist infinitely many vectors $v' \not\propto v$ that are observationally equivalent under linearization.
 
\paragraph{Proof Sketch. }
We establish non-identifiability via the null-space ambiguity argument. A related weight-space symmetry result is given in Appendix~\ref{app:gauge}.
 
\emph{Null-space ambiguity (primary mechanism).}
Under the local linear approximation (Equation~\ref{eq:first_order_expansion}), the steered output is
\begin{equation}
o(x, v, \alpha) \approx o_0(x) + \alpha J_\ell(x) v.
\end{equation}
 
For any $v_0 \in \ker(J_\ell)$, i.e., $J_\ell v_0 = 0$, define $v' = v + v_0$. Then
\begin{equation}
o(x, v', \alpha) \approx o_0(x) + \alpha J_\ell(x)(v + v_0) = o_0(x) + \alpha J_\ell(x) v \approx o(x, v, \alpha).
\end{equation}
 
Since $\dim(\ker(J_\ell)) = d - \mathrm{rank}(J_\ell)$, any case where $\mathrm{rank}(J_\ell) < d$ implies a non-trivial null space. In overparameterized language models the effective rank of $J_\ell$ is often smaller than $d$ \citep{oymak2019generalization}, yielding $\dim(\ker(J_\ell)) \ge 1$ in practice. Hence infinitely many such $v_0$ exist, producing observationally equivalent vectors $v' \not\propto v$.
 
\paragraph{Remark (Scope of linearization). }
Proposition~1 is proved under the local linear approximation (Equation~\ref{eq:first_order_expansion}), which holds exactly for piecewise-linear activations (ReLU MLPs on activation-region interiors) and as a first-order approximation for smooth nonlinearities (GeLU, attention softmax). For small steering strengths $\alpha$, the nonlinear residual is $O(\alpha^2)$ and empirically negligible. Extending the result to the exact nonlinear regime $o = F_{\ell \to L}(h_\ell + \alpha v)$ would require characterizing the Hessian structure of $F_{\ell \to L}$ and is left for future work.
 
\paragraph{Corollary 1.1 (Null-space equivalence). }
Under the linear approximation $o \approx o_0 + \alpha J_\ell v$, any vector $v' = v + v_0$ where $v_0 \in \ker(J_\ell)$ is observationally equivalent to $v$ since $J_\ell v' = J_\ell v + J_\ell v_0 = J_\ell v$.

\section{Empirical Validation}\label{sec:experiments}
 
We now validate that the non-identifiability characterized in Proposition~1 manifests in large language models. Our experiments test both the theoretical prediction directly — via null-space dimensionality estimation and orthogonal perturbation tests — and its broader consequences across extraction methods, steering strengths and prompt distributions. A complementary logit-level analysis in Appendix~\ref{app:logit} further examines output distribution preservation beyond the semantic probes used here.
 
\subsection{Setup}
 
We evaluate on Llama-3.1-8B-Instruct and Qwen2.5-3B-Instruct\footnote{For brevity, we refer to these models as Llama and Qwen respectively throughout.} \citep{yang2024qwen25, meta2024llama31} representing two architecture families. The steering interventions are applied at the middle layer of each model (layer 16/32 for Llama, layer 18/36 for Qwen) following standard practice \citep{chen2025persona}. We study five semantic traits: formality, politeness, sentiment, truthfulness and agreeableness which spans surface-level and cognitive attributes.
 
The trait presence in generated outputs is measured using pre-trained classifiers: formality with \texttt{s-nlp/roberta-base-formality-ranker}, sentiment with \texttt{cardiffnlp/twitter-roberta-base-sentiment-latest} and politeness, truthfulness and agreeableness with zero-shot classification via \texttt{facebook/bart-large-mnli}, all normalized to $[0, 1]$. For each trait, 50 contrastive prompt pairs $(x^+, x^-)$ are constructed to contrast high and low trait expression, and the steering vector is computed as the mean activation difference:
\begin{equation}
v = \frac{1}{50}\sum_{i=1}^{50}\bigl[h_\ell(x^+_i) - h_\ell(x^-_i)\bigr]
\end{equation}

For the vector equivalence experiment (Section~\ref{sec:vec_equiv}), a second vector is extracted via PCA of the activation covariance matrix over the same pairs. We report Cohen's $d$ as the primary metric throughout, using the threshold $|d| < 0.2$ (negligible effect) to indicate behavioral equivalence \citep{cohen2013statistical}.
 
\subsection{Orthogonal Perturbation Test}\label{sec:orthogonal}
 
We test whether adding a random direction orthogonal to $v$ changes steering behavior. For each trait and model, we generate $n$ random unit vectors $v_\perp$ orthogonal to $v$ via Gram-Schmidt and evaluate steering with $v + v_\perp$ versus $v$ alone across 100 held-out test prompts.
 
Results are shown in Table~\ref{tab:orthogonal}. Across all traits and both models, mean $|d| < 0.2$ for both $n = 5$ and $n = 10$ seeds. With $n = 10$ seeds, mean Cohen's $d$ is 0.119 for Llama and 0.131 for Qwen, well below the negligible-effect threshold. The effect sizes converge between $n=5$ and $n=10$ for all trait-model pairs, with confidence intervals tightening as sample size increases, indicating the observed equivalence is not a sampling artifact. Despite differences in architecture and scale, both models exhibit nearly identical patterns of non-identifiability, confirming the result is not model-specific.
 
\begin{table}[h]
\centering
\caption{\textbf{Orthogonal perturbation test}. Mean Cohen's $d$ (std) comparing $v$ and $v + v_\perp$ across $n$ seeds. All values below the equivalence threshold $|d| = 0.2$.}
\label{tab:orthogonal}
\small
\begin{tabular}{llcccc}
\toprule
\multirow{2}{*}{Trait} & \multirow{2}{*}{Model} & \multicolumn{2}{c}{$n=5$} & \multicolumn{2}{c}{$n=10$} \\
\cmidrule(lr){3-4}\cmidrule(lr){5-6}
 & & Mean & Std & Mean & Std \\
\midrule
\multirow{2}{*}{Formality}     & Llama & 0.120 & 0.037 & 0.059 & 0.056 \\
                                & Qwen  & 0.052 & 0.048 & 0.151 & 0.105 \\
\multirow{2}{*}{Politeness}    & Llama & 0.171 & 0.062 & 0.159 & 0.113 \\
                                & Qwen  & 0.065 & 0.032 & 0.120 & 0.091 \\
\multirow{2}{*}{Sentiment}     & Llama & 0.083 & 0.049 & 0.104 & 0.040 \\
                                & Qwen  & 0.206 & 0.141 & 0.150 & 0.079 \\
\multirow{2}{*}{Truthfulness}  & Llama & 0.045 & 0.027 & 0.085 & 0.070 \\
                                & Qwen  & 0.081 & 0.076 & 0.092 & 0.087 \\
\multirow{2}{*}{Agreeableness} & Llama & 0.159 & 0.059 & 0.186 & 0.067 \\
                                & Qwen  & 0.114 & 0.062 & 0.156 & 0.046 \\
\bottomrule
\end{tabular}
\end{table}

\subsection{Vector Equivalence Across Extraction Methods}\label{sec:vec_equiv}
 
We test whether non-identifiability extends beyond orthogonal perturbations to vectors extracted by different methods. We compare $v_1$ (contrastive mean difference) and $v_2$ (first principal component of activation variance) for each trait and model. These two methods are theoretically distinct: the contrastive method targets the mean activation shift across contrasting prompts while PCA targets the direction of maximum variance. If they produce equivalent behavioral effects despite geometric distinctness, this is direct evidence that the equivalence class is large enough to contain vectors from independent extraction pipelines.
 
\begin{figure}[h]
\centering

\begin{minipage}[t]{0.48\columnwidth}
    \centering
    \vspace{0pt}
    \includegraphics[width=\linewidth]{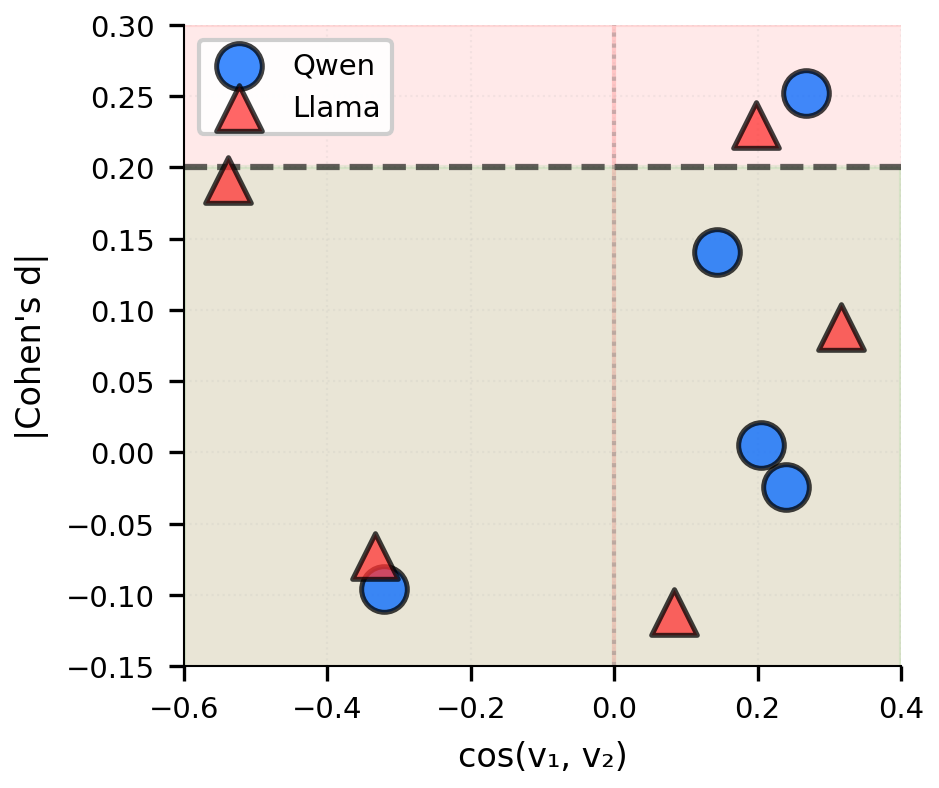}
    \captionof{figure}{\textbf{Geometric-semantic decoupling}. Despite large variation in cosine similarity between $v_1$ and $v_2$, almost all points fall below the equivalence threshold $|d| = 0.2$, proving that geometric diversity does not translate to behavioral difference.}
    \label{fig:vec_equiv}
\end{minipage}
\hfill
\begin{minipage}[t]{0.48\columnwidth}
    \centering
    \vspace{11pt} % moves table slightly down
    \small
    \setlength{\tabcolsep}{4pt}
    \begin{tabular}{llrr}
    \toprule
    Trait & Model & $\cos(v_1, v_2)$ & Cohen's $d$ \\
    \midrule
    \multirow{2}{*}{Agreeableness} & Qwen  & $+0.239$ & $-0.024$ \\
                                    & Llama & $-0.333$ & $-0.073$ \\
    \multirow{2}{*}{Formality}     & Qwen  & $+0.144$ & $+0.141$ \\
                                    & Llama & $+0.198$ & $+0.230$ \\
    \multirow{2}{*}{Politeness}    & Qwen  & $+0.267$ & $+0.252$ \\
                                    & Llama & $+0.317$ & $+0.088$ \\
    \multirow{2}{*}{Sentiment}     & Qwen  & $+0.205$ & $+0.005$ \\
                                    & Llama & $-0.539$ & $+0.191$ \\
    \multirow{2}{*}{Truthfulness}  & Qwen  & $-0.321$ & $-0.096$ \\
                                    & Llama & $+0.084$ & $-0.112$ \\
    \bottomrule
    \end{tabular}

    \vspace{28pt} % space between table and caption

    \captionof{table}{\textbf{Vector equivalence results}. $v_1$ = contrastive and $v_2$ = PCA. All Cohen's $d < 0.27$.}
    \label{tab:vec_equiv}
\end{minipage}

\end{figure}

Figure~\ref{fig:vec_equiv} shows the geometric-semantic decoupling: cosine similarities range from $-0.54$ to $+0.32$, indicating that $v_1$ and $v_2$ are often near-orthogonal or even anti-aligned. Despite this, Cohen's $d$ remains below $0.27$ in all conditions. The Llama sentiment pair ($\cos = -0.54$, $|d| = 0.19$) is the most striking: two vectors pointing in nearly opposite directions produce statistically equivalent behavioral effects, confirming that the extraction method does not determine a unique steering direction. Full numerical results are provided in Table~\ref{tab:vec_equiv}.
 
\subsection{Scale Invariance}\label{sec:alpha_sweep}
 
\begin{figure}[h]
    \centering
    \includegraphics[width=\linewidth]{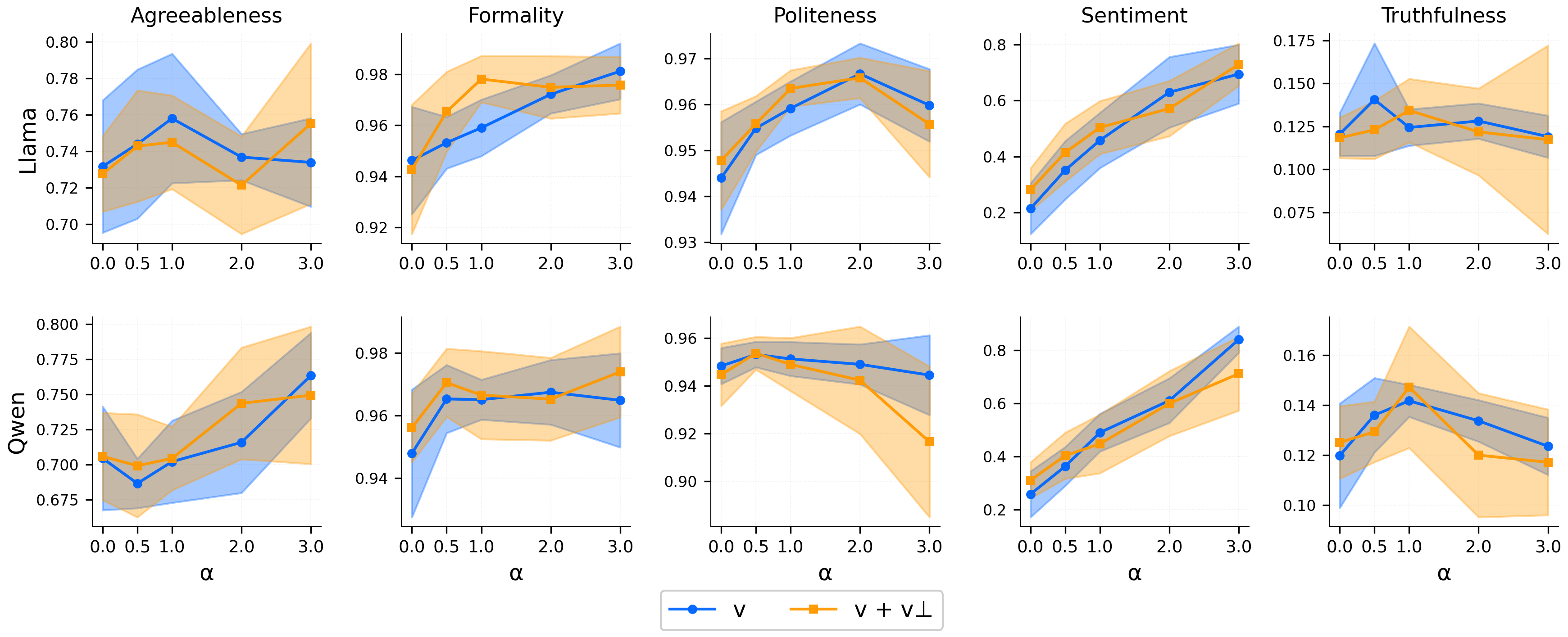}
    \caption{\textbf{Semantic equivalence across operational steering range} ($\alpha \in [0.0, 3.0]$). Blue ($v$) and orange ($v + v_{\perp}$) trajectories track classifier scores, with shaded confidence bands. Near-overlapping curves confirm non-identifiability throughout standard steering regime.}
    \label{fig:alpha_sweep}
\end{figure}

Proposition~1 predicts that non-identifiability holds throughout the regime where the linear approximation is valid. We test this by sweeping steering strength $\alpha \in \{0.0, 0.5, 1.0, 2.0, 3.0\}$ and measuring Cohen's $d$ between $v$ and $v + v_\perp$ at each value, across 10 random seeds. Figure~\ref{fig:alpha_sweep} shows results for all five traits and both models. 

The semantic scores for $v$ and $v + v_\perp$ track closely throughout $\alpha \leq 3$, with overlapping confidence bands in nearly all conditions. Sentiment exhibits wider variance due to the binary nature of its classifier output but remains well below the equivalence threshold on average.
 
\subsection{Null-Space Spanning}\label{sec:spanning}
 
The orthogonal perturbation test uses a single random direction per seed. A stronger test is whether equivalence holds across the full null space: if 50 independently sampled orthogonal directions all produce small effects, the equivalence class is high-dimensional rather than confined to a thin region around $v$. We sample 50 random unit vectors orthogonal to $v$ via Gram-Schmidt and measure Cohen's $d$ for each direction individually at $\alpha = 1.0$.
 
\begin{figure}[h]
\centering

\begin{minipage}[t]{0.48\columnwidth}
    \centering
    \vspace{0pt}
    \includegraphics[width=\linewidth]{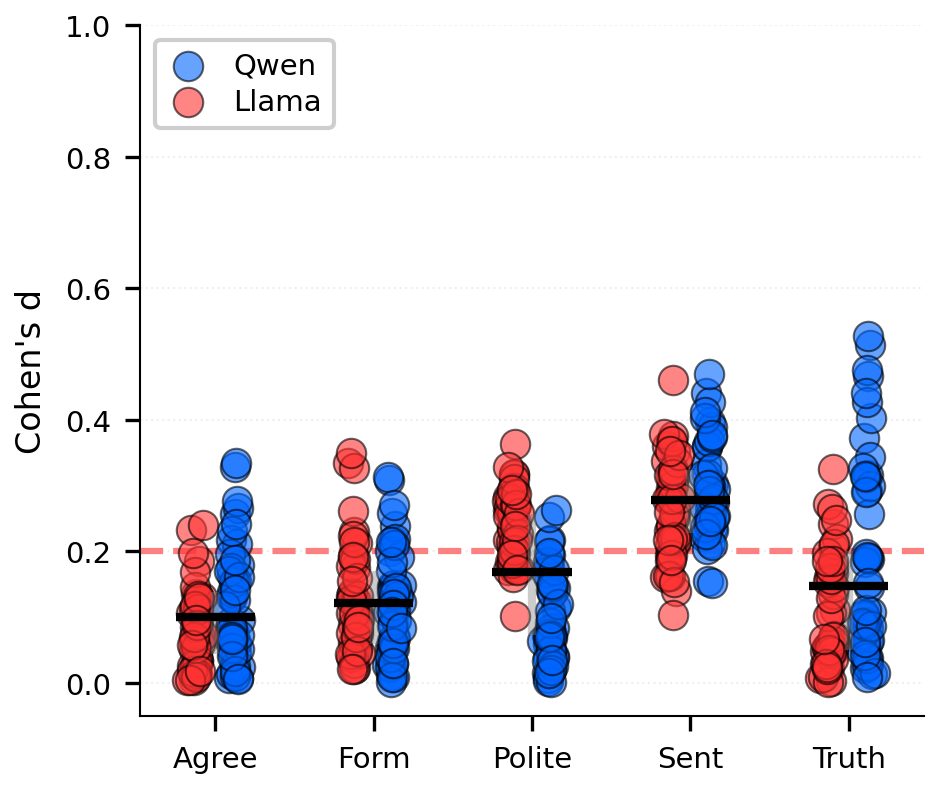}

    \vspace{6pt}

    \captionof{figure}{Cohen's $d$ for 50 random orthogonal directions per trait across both models (Qwen: blue, Llama: red).}
    \label{fig:spanning}
\end{minipage}
\hfill
\begin{minipage}[t]{0.48\columnwidth}
    \centering
    \vspace{2pt}
    \small
    \renewcommand{\arraystretch}{1.2} % adjust (e.g., 1.1–1.5)

    \begin{tabular}{llc}
    \toprule
    \textbf{Trait} & \textbf{Model} & \textbf{Mean $|d|$} \\
    \midrule
    \multirow{2}{*}{Agreeableness} & Qwen  & 0.121 \\
                                    & Llama & 0.080 \\
    \multirow{2}{*}{Formality}     & Qwen  & 0.119 \\
                                    & Llama & 0.124 \\
    \multirow{2}{*}{Politeness}    & Qwen  & 0.095 \\
                                    & Llama & 0.243 \\
    \multirow{2}{*}{Sentiment}     & Qwen  & 0.291 \\
                                    & Llama & 0.265 \\
    \multirow{2}{*}{Truthfulness}  & Qwen  & 0.185 \\
                                    & Llama & 0.108 \\
    \bottomrule
    \end{tabular}

    \vspace{20pt}

    \captionof{table}{\textbf{Null-space spanning results}. Mean Cohen's $d$ across 50 random orthogonal directions per trait at $\alpha = 1.0$.}
    \label{tab:nullspace_span}
\end{minipage}

\end{figure}

Figure~\ref{fig:spanning} and Table~\ref{tab:nullspace_span} show results for all traits and both models. Most traits cluster near $|d| \approx 0.08$--$0.13$, with nearly all 50 directions per trait falling below the equivalence threshold ($|d| = 0.2$). Sentiment and politeness show slightly elevated responses ($|d| \approx 0.24$--$0.29$), though these remain well within the small-effect range and do not indicate a breakdown of equivalence. The key result is that equivalence is not isolated to a few directions: it is the dominant behavior across the null space, confirming that the equivalence class is genuinely high-dimensional.
 
\subsection{Multi-Environment Validation}\label{sec:multi_env}
 
\begin{figure}[h]
    \centering
    \includegraphics[width=\columnwidth]{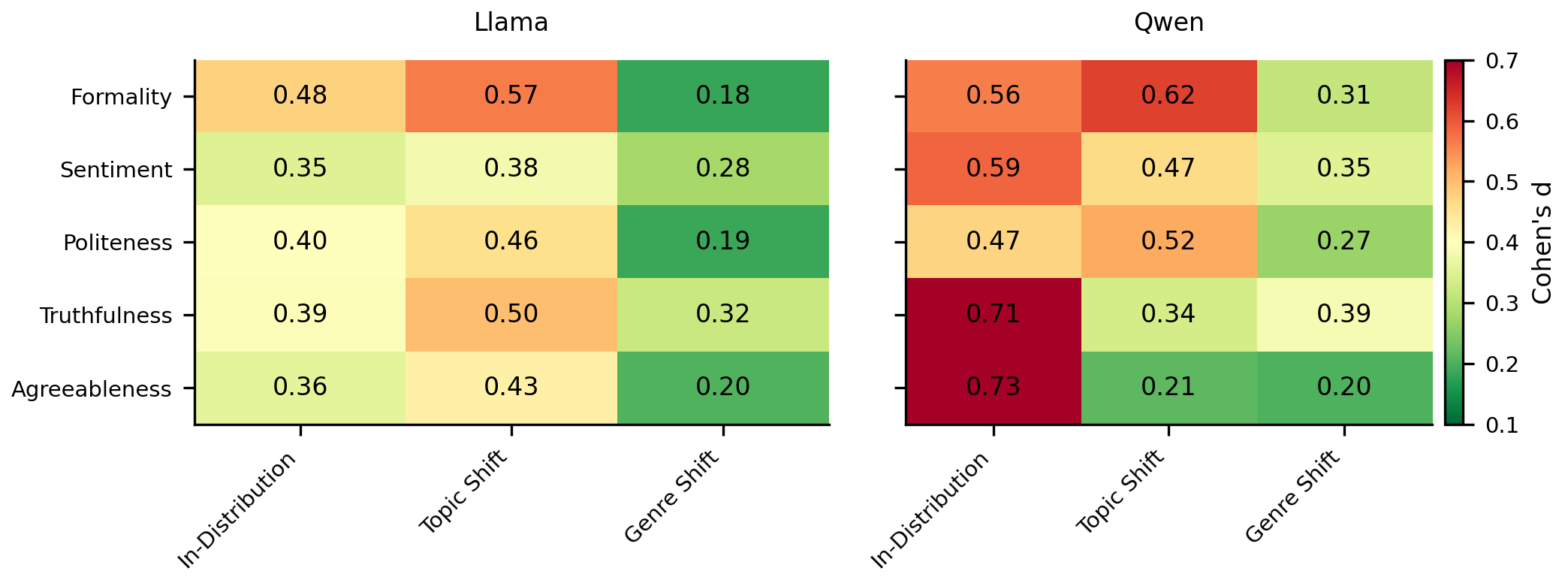}
    \caption{Cohen's $d$ for $v$ vs.\ $v + v_\perp$ across the three prompt environments. Color gradient represents effect magnitude from small to large across traits and models.}
    \label{fig:multi_env}
\end{figure}

To confirm that non-identifiability is not an artifact of limited prompt diversity, we test whether equivalence persists across distribution shift. We extract a steering vector from in-distribution prompts and evaluate on three environments: in-distribution, topic-shift, and genre-shift. Cohen's $d$ is measured between $v$-steered and $(v + v_\perp)$-steered outputs in each environment. The results are shown in Figure~\ref{fig:multi_env}. 

Effect sizes remain moderate across all environments and traits, with most conditions in the small-effect range. No catastrophic breakdown occurs under topic or genre shift, confirming that the equivalence class structure is a stable geometric property of the model rather than a prompt-specific artifact. The elevated values for truthfulness and agreeableness under in-distribution steering ($d \approx 0.7$) reflect higher classifier variance for these cognitively complex traits rather than a genuine breakdown of equivalence, consistent with findings that abstract semantic properties are harder to probe reliably via linear classifiers \citep{elazar2021amnesic, marks2023geometry}.

\section{Discussion}\label{sec:discussion}
 
Our results establish that steering vectors are non-identifiable in a strong empirical sense: orthogonal perturbations, alternative extraction methods and diverse prompt environments all fail to distinguish between behaviorally equivalent directions. Claims that a steering vector \emph{represents} a specific semantic concept are therefore not warranted by behavioral evidence alone. Since many geometrically distinct vectors produce identical outputs, the observed behavioral effect of steering does not localize semantic content to a specific direction in activation space. We note that this does not mean steering is ineffective: the experiments confirm that steering vectors reliably shift model behavior in the intended direction. The limitation is representational, not functional.
 
\section{Limitations}\label{sec:limitations}
 
Our empirical validation covers two models at mid-network layers across five semantic traits. While our theoretical results apply generally, the empirical magnitude of non-identifiability may vary across model families, scales, architectures and layer positions. The multi-environment experiments demonstrate that equivalence persists across diverse prompt distributions but do not test whether explicit structural interventions such as independence constraints \citep{hyvarinen1999nonlinear}, sparsity regularization or invariance objectives \citep{arjovsky2019invariant} can recover identifiable representations. We also do not directly verify that the semantic null-space estimated via activation SVD coincides with the Jacobian null-space assumed in Proposition~1; this connection is supported by the behavioral results but not formally proven.
 
\section{Conclusion}\label{sec:conclusion}
 
We have shown that steering vectors are fundamentally non-identifiable under white-box single-layer access: infinitely many geometrically distinct directions produce equivalent behavioral effects due to the large null space of the activation-to-output Jacobian. This non-identifiability persists across models, traits, extraction methods and prompt distributions. The implication is that behavioral equivalence is not a sufficient basis for interpretability claims about internal representations, and that alignment methods relying on steering vectors require structural constraints beyond behavioral testing to be trusted. More broadly, any intervention-based interpretability method that validates representational claims solely through output behavior faces the same fundamental limitation.

\section*{Acknowledgments}
We thank Vast.ai for GPU compute resources. We acknowledge the HuggingFace platform and maintainers of the open-source models used in this work.
 
\section*{Ethics Statement}
This work studies the theoretical and empirical foundations of activation steering in large language models. We do not introduce new steering methods or systems that could enable misuse. Our findings only highlight limitations of existing methods. We use publicly available models and standard evaluation datasets throughout and have adhered to their respective terms of use.

\section*{Disclosure of AI Usage}
Large language models were used for language editing of the draft and code assistance in this work. All research ideas, technical content, experimental design, results and conclusions are the original work of the authors.

\bibliography{colm2026_conference}
\bibliographystyle{colm2026_conference}

\appendix

\section{Null-Space Dimensionality}
\label{app:nullspace_dim}

We estimate the dimensionality of the semantically inert subspace at three layers (L/4, L/2, 3L/4) of each model. Hidden activations are collected over 200 in-distribution prompts; we compute the activation covariance matrix and apply SVD. Effective rank is defined as the number of singular values exceeding $\varepsilon = 0.01$. The null-space fraction is $(d - \text{rank}) / d$.
 
\begin{table}[h]
\centering
\caption{Null-space dimensionality across three layers for both models.}
\label{tab:nullspace_app}
\small

\begin{tabular}{llrrr}
\toprule
\multirow{2}{*}{Model} & \multirow{2}{*}{Layer} & \multicolumn{3}{c}{Metrics} \\
\cmidrule(lr){3-5}
 & & $d$ & Eff.\ rank & NF \\
\midrule
\multirow{3}{*}{Llama} & L/4   & 4096 & 57  & 0.943 \\
                              & L/2  & 4096 & 59  & 0.941 \\
                              & 3L/4 & 4096 & 63  & 0.937 \\
\multirow{3}{*}{Qwen}   & L/4  & 2048 & 139 & 0.861 \\
                              & L/2  & 2048 & 127 & 0.873 \\
                              & 3L/4 & 2048 & 116 & 0.884 \\
\bottomrule
\end{tabular}
\end{table}
 
\begin{figure}[h]
\centering

\begin{subfigure}[t]{0.48\columnwidth}
    \centering
    \includegraphics[width=\linewidth]{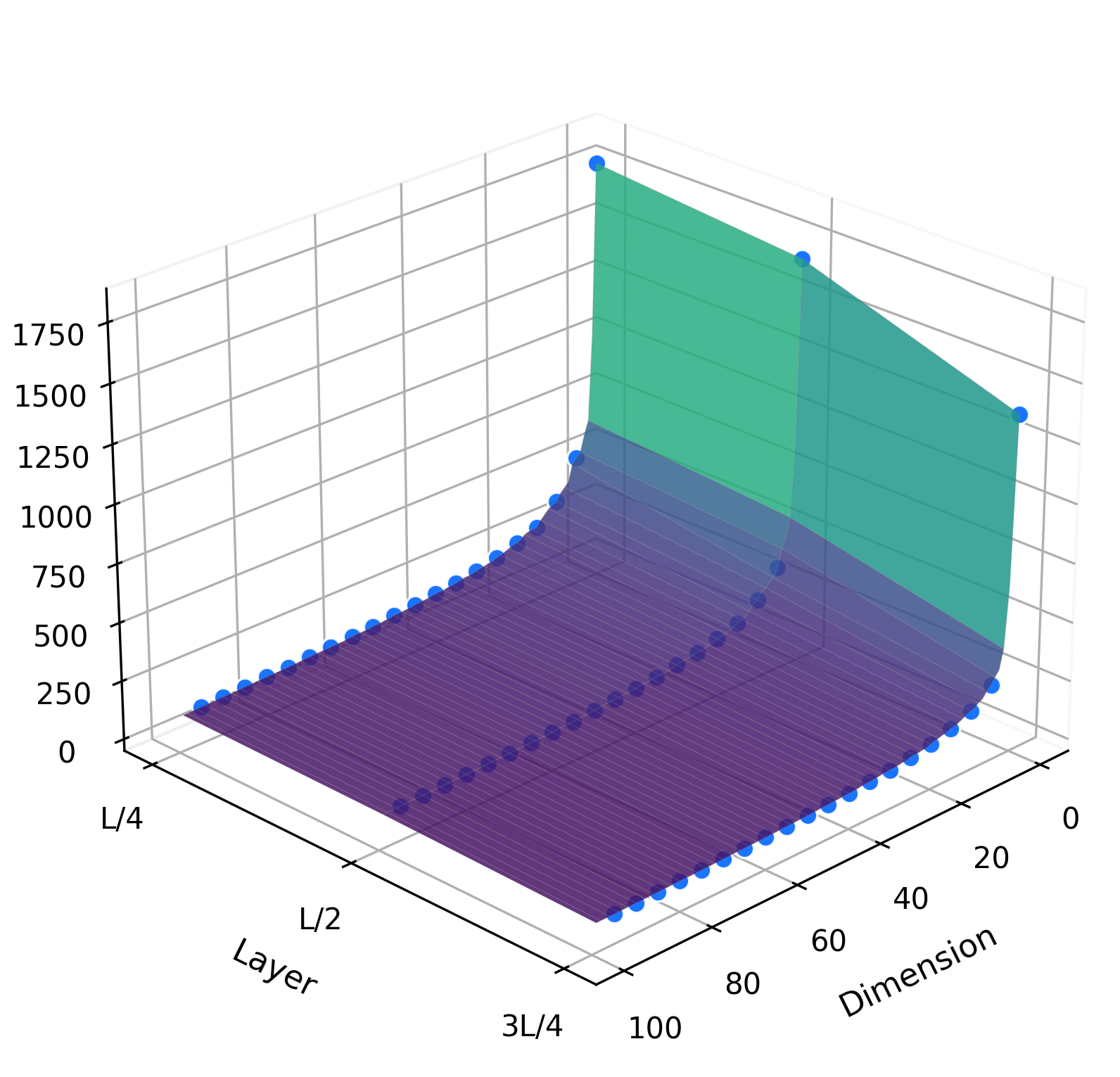}
    \caption{}
    \label{fig:nullspace_qwen}
\end{subfigure}
\hfill
\begin{subfigure}[t]{0.48\columnwidth}
    \centering
    \includegraphics[width=\linewidth]{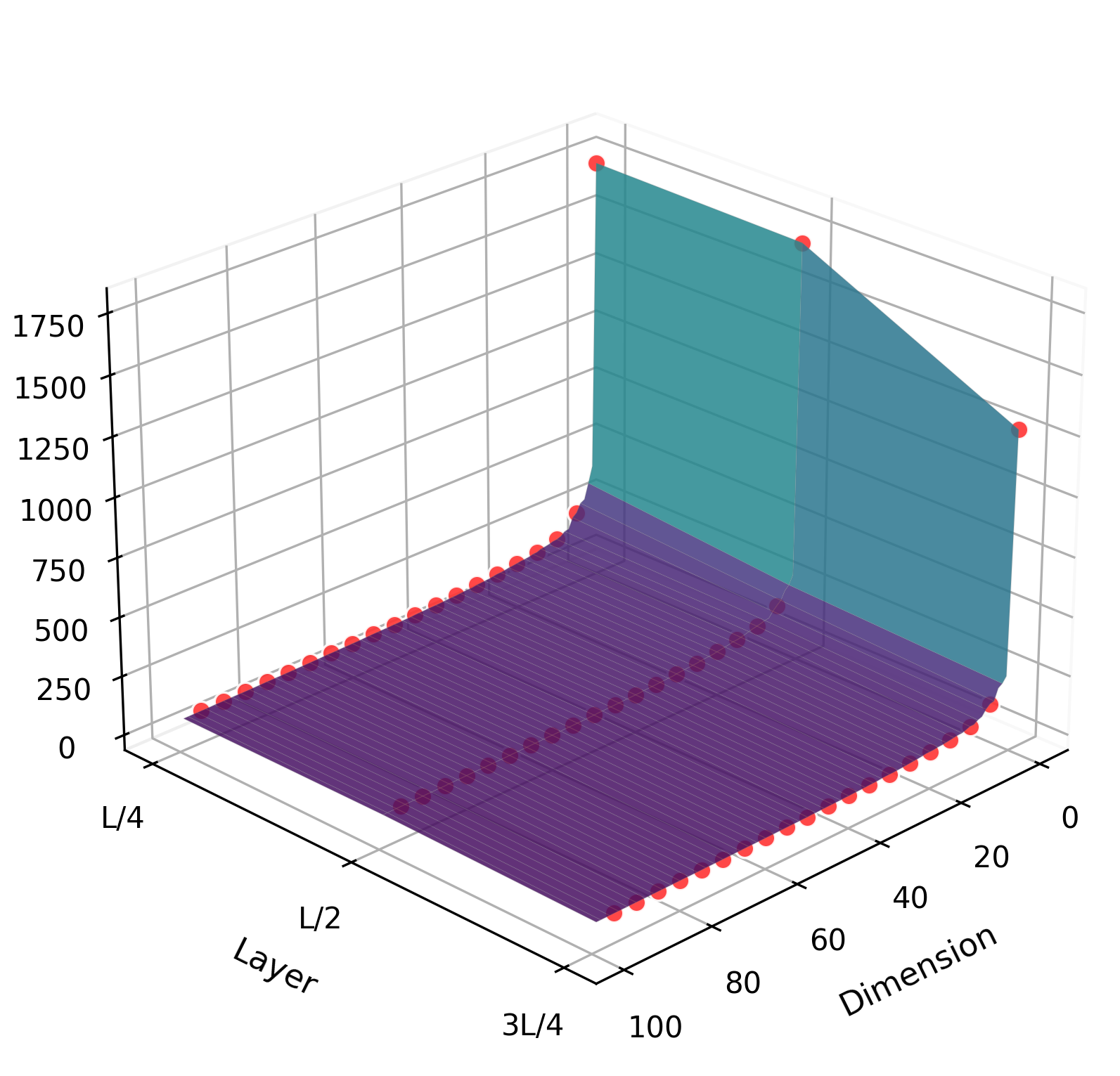}
    \caption{}
    \label{fig:nullspace_llama}
\end{subfigure}

\caption{Singular value spectra across layers. (a) Qwen and (b) Llama both exhibit a sharp cliff after the first 50--140 dimensions followed by a near-zero plateau, confirming that the vast majority of the activation space is semantically inert across all layers.}
\end{figure}

The singular value spectra at three layers are shown in Figures~\ref{fig:nullspace_qwen} and~\ref{fig:nullspace_llama}. The surface drops sharply after the first few dimensions, with singular values collapsing to near zero beyond dimension 50--140 depending on the model. This cliff-like decay persists consistently across all three layers, indicating that the low-rank structure is not a property of any particular depth but a stable geometric feature of both architectures. The vast flat region beyond the initial peak represents the semantically inert null space: directions in this region produce no measurable change in output logits and are therefore undetectable by any behavioral evaluation. For Qwen, the decay is steeper and the plateau lower which is consistent with its smaller hidden dimension ($d = 2048$) relative to Llama ($d = 4096$).
 
\section{Logit-Level Equivalence Test}
\label{app:logit}
% ------------------------------------------------------------
 
To confirm that equivalence is not an artifact of semantic classifier
choice, we measure raw logit distances. For each model and trait, we
compute the $\ell_2$ distance between the next-token logit vectors
produced under three conditions: (i) steering with $v$, (ii) steering
with $v + v_\perp$ where $v_\perp$ is a norm-matched orthogonal
direction, and (iii) steering with a norm-matched random direction
$v_\text{rand}$. The ratio
\begin{equation}
R = \frac{\|\ell(v) - \ell(v + v_\perp)\|_2}
         {\|\ell(v) - \ell(v_\text{rand})\|_2}
\end{equation}
quantifies how much smaller the logit shift from an orthogonal
perturbation is compared to a random one. Values $R < 1$ indicate that
orthogonal perturbations are systematically less disruptive than random
directions.
 
\begin{table}[h]
\centering
\caption{Logit-level equivalence. Baseline = $\|\ell(0) - \ell(v)\|_2$; $v_\perp$ and Random denote $\|\ell(v) - \ell(v + v_\perp)\|_2$ and $\|\ell(v) - \ell(v_\text{rand})\|_2$ respectively. $R < 1$ in all conditions.}
\label{tab:logit}
\small

\begin{tabular}{llrrrr}
\toprule
\multirow{2}{*}{Model} & \multirow{2}{*}{Trait} & \multicolumn{4}{c}{Metrics} \\
\cmidrule(lr){3-6}
 & & Baseline & $v_\perp$ & Random & $R$ \\
\midrule
\multirow{5}{*}{Qwen}
  & Formality     & 43.73  & 40.93  & 57.78  & 0.71 \\
  & Sentiment     & 67.56  & 62.79  & 86.21  & 0.73 \\
  & Truthfulness  & 113.06 & 118.24 & 165.91 & 0.71 \\
  & Politeness    & 53.14  & 52.95  & 73.07  & 0.72 \\
  & Agreeableness & 75.66  & 64.64  & 96.92  & 0.67 \\
\cmidrule(lr){1-6}
\multirow{5}{*}{Llama}
  & Formality     & 119.12 & 64.36  & 136.58 & 0.47 \\
  & Sentiment     & 87.67  & 61.84  & 100.94 & 0.61 \\
  & Truthfulness  & 206.04 & 121.33 & 226.73 & 0.54 \\
  & Politeness    & 116.88 & 98.70  & 144.28 & 0.68 \\
  & Agreeableness & 119.98 & 85.01  & 128.02 & 0.66 \\
\bottomrule
\end{tabular}

\end{table}
 
\begin{figure}[h]
\centering

\begin{subfigure}[t]{0.48\columnwidth}
    \centering
    \includegraphics[width=\linewidth]{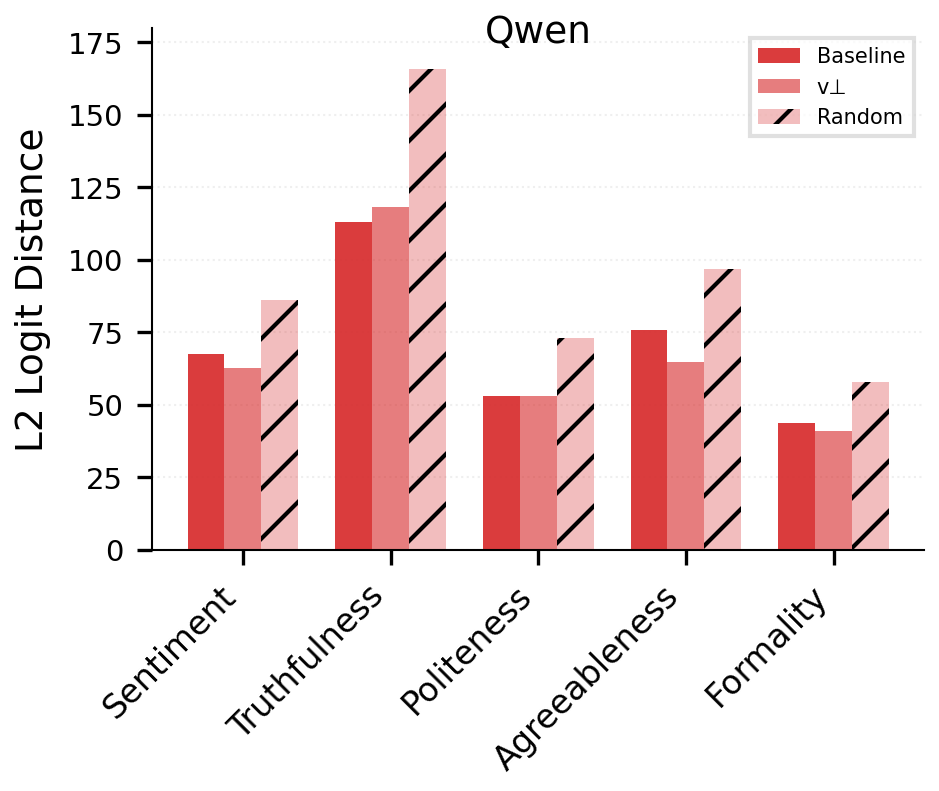}
    \caption{}
    \label{fig:logit_qwen}
\end{subfigure}
\hfill
\begin{subfigure}[t]{0.48\columnwidth}
    \centering
    \includegraphics[width=\linewidth]{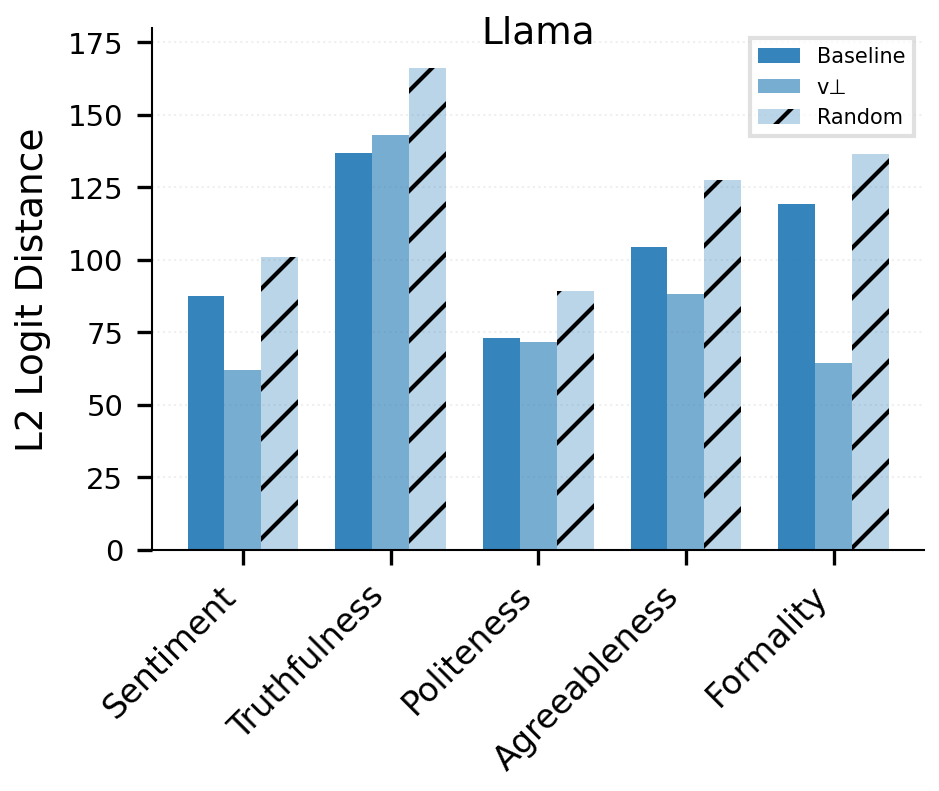}
    \caption{}
    \label{fig:logit_llama}
\end{subfigure}

\caption{Logit $\ell_2$ distances for (a) Qwen and (b) Llama. Each group shows three bars per trait: Baseline ($\|\ell(0) - \ell(v)\|_2$), $v_\perp$ ($\|\ell(v) - \ell(v+v_\perp)\|_2$) and Random ($\|\ell(v) - \ell(v_\text{rand})\|_2$). The $v_\perp$ bar is consistently shorter than Random across all traits which confirms that orthogonal perturbations are less disruptive to the output distribution than unstructured directions of equal norm.}
\end{figure}

$R$ ranges from $0.47$ to $0.73$ across all conditions, as reported in Table~\ref{tab:logit}. Orthogonal perturbations induce logit shifts that are 27--53\% smaller than random directions of the same norm. In several Llama conditions, the orthogonal perturbation logit distance is also smaller than the baseline distance, indicating that $v_\perp$ moves the model less than $v$ itself does. These results confirm that the equivalence observed in semantic scores is not a property of the classifiers: it is present directly in the model's output distribution. The full bar-chart comparisons are shown in Figures~\ref{fig:logit_qwen} and~\ref{fig:logit_llama}.
 
\section{Detailed Proof of Proposition 1}
\label{app:proof}
% ------------------------------------------------------------
 
We provide a constructive proof of Proposition~1, formalizing the
null-space ambiguity argument presented in Section~\ref{sec:results}.
 
\paragraph{Setup. } Consider the local linear approximation of the steering effect at layer
$\ell$:
\begin{equation}
o(x, v, \alpha) \approx o(x, 0, 0) + \alpha J_\ell(x) v
\end{equation}
where $J_\ell(x) = \frac{\partial o}{\partial h_\ell}\big|_{h_\ell(x)}
\in \mathbb{R}^{V \times d}$ is the Jacobian of the output logits with
respect to the layer-$\ell$ activations, $V$ is the vocabulary size, and
$d$ is the hidden dimension.
 
\subsection{Null-space characterization}
 
Define the null space of $J_\ell$ as:
\begin{equation}
\mathcal{N} = \ker(J_\ell) = \{v_0 \in \mathbb{R}^d : J_\ell v_0 = 0\}
\end{equation}
By the rank-nullity theorem:
\begin{equation}
\dim(\mathcal{N}) = d - \operatorname{rank}(J_\ell)
\end{equation}
 
\subsection{Rank bound and effective rank}
 
The Jacobian $J_\ell \in \mathbb{R}^{V \times d}$ satisfies
$\operatorname{rank}(J_\ell) \leq \min(V, d)$. In modern language models
$V \approx 50{,}000$ and $d \approx 2{,}000$--$8{,}000$, so the
theoretical maximum rank is $d$. However, empirical evidence indicates
that output distributions lie on a low-dimensional manifold. The
effective rank, defined as
\begin{equation}
\operatorname{rank}_\varepsilon(J_\ell) =
\#\{\sigma_i : \sigma_i > \varepsilon \cdot \sigma_{\max}\}
\end{equation}
where $\sigma_i$ are singular values of $J_\ell$ and $\varepsilon$ is a
threshold (e.g., $10^{-2}$), is substantially lower than $d$ in
practice. Our measurements (Appendix~\ref{app:nullspace_dim}) yield
effective ranks of 57 and 139 for Llama and Qwen respectively, leaving
null-space fractions of 94\% and 86\%. The constructive argument below
requires only $\dim(\mathcal{N}) \geq 1$.
 
\subsection{Constructing observationally equivalent vectors}
 
For any steering vector $v \in \mathbb{R}^d$ and any
$v_0 \in \mathcal{N}$, define $v' = v + v_0$. Then for all prompts $x$
and all steering strengths $\alpha$:
\begin{equation}
J_\ell(x) v' = J_\ell(x)(v + v_0) = J_\ell(x) v + J_\ell(x) v_0 = J_\ell(x) v
\end{equation}

where the last step uses $v_0 \in \ker(J_\ell)$. Therefore:
\begin{equation}
o(x, v', \alpha) \approx o(x, 0, 0) + \alpha J_\ell(x) v' = o(x, 0, 0) + \alpha J_\ell(x) v \approx o(x, v, \alpha)
\end{equation}
 
\subsection{Infinitely many distinct solutions}
 
Since $\dim(\mathcal{N}) \geq 1$, the null space contains infinitely
many directions. For any $v_0 \in \mathcal{N} \setminus \{0\}$ and any
$\beta \in \mathbb{R}$, the family
\begin{equation}
v'_\beta = v + \beta v_0
\end{equation}
generates infinitely many observationally equivalent vectors. To ensure
$v'_\beta \not\propto v$, we need $v + \beta v_0 \neq c v$ for any
scalar $c$, i.e., $\beta v_0 \neq (c - 1) v$, which fails only when
$v \in \operatorname{span}(v_0)$. For generic $v$ and
$v_0 \in \mathcal{N}$, this condition holds with probability zero.
Therefore, for almost all $\beta \neq 0$, the vector $v'_\beta$ is
geometrically distinct from $v$ yet observationally equivalent.
 
\paragraph{Conclusion.} Under the local linear approximation and whenever
$\operatorname{rank}(J_\ell) < d$, there exist infinitely many
geometrically distinct steering vectors that are observationally
equivalent to $v$. Proposition~1 follows. $\square$
 
\subsection{Scope of the linearization}

The proof relies on the approximation in
Equation~\ref{eq:first_order_expansion}. This holds exactly for
piecewise-linear activations (e.g., ReLU MLPs, on activation-region
interiors) and as a first-order approximation for smooth nonlinearities
(GeLU, attention softmax) with error $O(\alpha^2)$. For the steering
strengths used in our experiments ($\alpha \leq 3$), the nonlinear
residual is empirically negligible: the scale-invariance results in
Section~\ref{sec:alpha_sweep} confirm that equivalence holds across the
full operationally relevant range without degradation. Extending the
proof to the exact nonlinear regime would require characterizing the
Hessian structure of $F_{\ell \to L}$ and is left for future work.

\section{Weight-Space Gauge Symmetry}
\label{app:gauge}
% ------------------------------------------------------------
 
This section records a supplementary observation about gauge symmetry in neural network parameterization. It is \emph{distinct} from the null-space argument of Proposition~1: rather than fixing model weights and asking whether $v$ is recoverable from behavioral observations, we ask whether different weight-vector pairs $(W, v)$ and $(W', v')$ can produce identical input--output behavior. The answer is yes, via the linear reparameterization argument below. This symmetry is not realizable as a steering operation on a frozen model since it requires modifying $W_{\ell+1}$. We include it because it establishes a complementary non-identifiability result at the level of model parameterization.
 
\paragraph{Representation reparameterization.} Consider an invertible transformation $T: \mathbb{R}^d \to \mathbb{R}^d$ and define the reparameterized representation $h'_\ell(x) = T(h_\ell(x))$. If the subsequent layers satisfy $F_{\ell \to L}(h_\ell) = F'_{\ell \to L}(T(h_\ell))$, then the pair $(h_\ell, v)$ and $(h'_\ell, v')$ with $v' = DT(h_\ell) \cdot v$ are observationally indistinguishable. We now show this holds for linear $T$.
 
\paragraph{Linear reparameterization.} Restrict to $T(h) = Ah$ where $A \in \mathbb{R}^{d \times d}$ is invertible, so $h'_\ell(x) = Ah_\ell(x)$. For layer $\ell+1$ with weight matrix $W_{\ell+1}$ and bias $b_{\ell+1}$, define the reparameterized weights $W'_{\ell+1} = W_{\ell+1}A^{-1}$ and $b'_{\ell+1} = b_{\ell+1}$. The bias is invariant because we restrict to origin-preserving transformations; for affine $T(h) = Ah + c$ the bias transforms as $b'_{\ell+1} = b_{\ell+1} - W_{\ell+1}A^{-1}c$, but linear $T$ suffices to establish non-identifiability. Under this reparameterization:
\begin{equation}
h'_{\ell+1} = \sigma(W'_{\ell+1} h'_\ell + b'_{\ell+1}) = \sigma(W_{\ell+1}A^{-1} A h_\ell + b_{\ell+1}) = \sigma(W_{\ell+1} h_\ell + b_{\ell+1}) = h_{\ell+1}
\end{equation}
so the reparameterized network produces identical activations at layer $\ell+1$ and all subsequent layers.
 
\paragraph{Steering under reparameterization.} Under the original parameterization, steering applies $\tilde{h}_\ell = h_\ell + \alpha v$, giving $\tilde{h}_{\ell+1} = \sigma(W_{\ell+1}(h_\ell + \alpha v) + b_{\ell+1})$. Under the reparameterization with $v' = Av$, the steered representation at layer $\ell$ is:
\begin{equation}
\tilde{h}'_\ell = h'_\ell + \alpha v' = Ah_\ell + \alpha Av = A(h_\ell + \alpha v) = A\tilde{h}_\ell
\end{equation}
and the next-layer activation is:
\begin{equation}
\tilde{h}'_{\ell+1} = \sigma\!\left(W'_{\ell+1}(h'_\ell + \alpha v') + b'_{\ell+1}\right) = \sigma\!\left(W_{\ell+1}A^{-1}A(h_\ell + \alpha v) + b_{\ell+1}\right) = \tilde{h}_{\ell+1}
\end{equation}
so the steered activation at layer $\ell+1$ is identical under both parameterizations.
 
\paragraph{Infinitely many reparameterizations.} For any invertible $A$ with $A \neq cI$, the steering vectors $v' = Av$ and $v$ are geometrically distinct ($v' \not\propto v$) yet produce identical outputs. The space of invertible matrices with $A \neq cI$ has dimension $d^2 - 1$, providing infinitely many such reparameterizations. Non-identifiability therefore persists even when model weights are not fixed: infinitely many $(W, v)$ and $(W', v')$ pairs yield identical input--output behavior, confirming that the ambiguity is a structural property of the parameterization rather than a limitation of any particular extraction method.

\section{Mathematical Derivations}
\label{app:math}
% ------------------------------------------------------------
 
\subsection{Dimension of the Observational Equivalence Class}
 
For a steering vector $v \in \mathbb{R}^d$ and Jacobian $J_\ell \in \mathbb{R}^{V \times d}$ with rank $r$, two vectors $v$ and $v'$ are observationally equivalent if and only if $J_\ell v = J_\ell v'$, i.e., $J_\ell(v' - v) = 0$, i.e., $v' - v \in \ker(J_\ell)$. The equivalence class is therefore the affine subspace:
\begin{equation}
[v] = \{v' \in \mathbb{R}^d : J_\ell v' = J_\ell v\} = v + \ker(J_\ell)
\end{equation}
 
By the rank-nullity theorem, $\dim(\ker(J_\ell)) = d - r$. Let $\{u_1, \ldots, u_{d-r}\}$ be an orthonormal basis for $\ker(J_\ell)$ obtained via SVD of $J_\ell$. Then:
\begin{equation}
[v] = \left\{v + \sum_{i=1}^{d-r} \beta_i u_i : \beta_i \in \mathbb{R}\right\}
\end{equation}
The measure of this set in $\mathbb{R}^d$ is:
\begin{equation}
\text{vol}([v]) = \int_{\mathbb{R}^{d-r}} d\beta_1 \cdots d\beta_{d-r} = \infty
\end{equation}
so the equivalence class has infinite volume in $\mathbb{R}^d$ whenever $r < d$. Additionally, $v$ and $cv$ for $c \neq 0$ produce equivalent outputs up to rescaling $\alpha \to c\alpha$, so the equivalence class modulo scaling is a $(d-r)$-dimensional projective space $\mathbb{P}^{d-r}$. The practical implication is that steering vector recovery is a severely underdetermined inverse problem: there are infinitely many solutions consistent with any finite set of behavioral observations, parameterized by the free coefficients $\beta_1, \ldots, \beta_{d-r}$.
 
To quantify the degree of under-determination, define the \emph{non-identifiability ratio}:
\begin{equation}
\rho = \frac{d - r}{r} = \frac{\dim(\ker(J_\ell))}{\operatorname{rank}(J_\ell)}
\end{equation}
which measures unidentifiable directions per identifiable one. For Llama-3.1-8B with $d = 4096$ and $r = 57$:
\begin{equation}
\rho_\text{Llama} = \frac{4096 - 57}{57} = \frac{4039}{57} \approx 70.9
\end{equation}
For Qwen2.5-3B with $d = 2048$ and $r = 139$:
\begin{equation}
\rho_\text{Qwen} = \frac{2048 - 139}{139} = \frac{1909}{139} \approx 13.7
\end{equation}
In both cases $\rho \gg 1$: the unobservable dimensions vastly outnumber the observable ones. The null space is not a thin subspace but the dominant structure of the ambient geometry.
 
\subsection{Fisher Information and Cramér-Rao Bound}
 
We formalize why collecting more behavioral data cannot resolve the null-space ambiguity. Consider the linear Gaussian observation model for a single prompt $x$:
\begin{equation}
o(x) = J_\ell(x) v + \eta, \quad \eta \sim \mathcal{N}(0, \sigma^2 I_V)
\end{equation}
where $o(x) \in \mathbb{R}^V$ are the observed output logits and $v \in \mathbb{R}^d$ is the parameter to be estimated. The log-likelihood is:
\begin{equation}
\log p(o \mid x, v) = -\frac{1}{2\sigma^2}\|o - J_\ell(x) v\|^2 + \text{const}
\end{equation}
The score function is $\frac{\partial \log p}{\partial v} = \frac{1}{\sigma^2} J_\ell(x)^\top (o - J_\ell(x) v)$, and the Fisher information matrix is:
\begin{equation}
\mathcal{I}(v) = \mathbb{E}_o\left[\frac{\partial \log p}{\partial v}\frac{\partial \log p}{\partial v}^\top\right] = \frac{1}{\sigma^2} J_\ell(x)^\top J_\ell(x)
\end{equation}
 
The Cramér-Rao lower bound states that for any unbiased estimator $\hat{v}$:
\begin{equation}
\text{Cov}(\hat{v}) \succeq \mathcal{I}(v)^{-1} = \sigma^2 \bigl(J_\ell(x)^\top J_\ell(x)\bigr)^+
\end{equation}
where $(\cdot)^+$ is the Moore-Penrose pseudoinverse. For a null-space direction $u_i \in \ker(J_\ell)$, we have $J_\ell u_i = 0$, so:
\begin{equation}
u_i^\top \mathcal{I}(v) u_i = \frac{1}{\sigma^2} u_i^\top J_\ell^\top J_\ell u_i = \frac{1}{\sigma^2}\|J_\ell u_i\|^2 = 0
\end{equation}
Since $\mathcal{I}(v)$ is singular in the direction $u_i$, its pseudoinverse satisfies $u_i^\top \mathcal{I}(v)^+ u_i = \infty$ (the pseudoinverse assigns zero eigenvalue to $u_i$, meaning the bound is vacuously infinite). Therefore:
\begin{equation}
\text{Var}(u_i^\top \hat{v}) \geq u_i^\top \mathcal{I}(v)^+ u_i = \infty \quad \forall\, u_i \in \ker(J_\ell)
\end{equation}
 
\paragraph{Multi-prompt extension.} Stacking observations across $N$ prompts $\{x_1, \ldots, x_N\}$ gives:
\begin{equation}
\mathbf{o} = J_\text{stack} v + \boldsymbol{\eta}, \quad J_\text{stack} = \begin{bmatrix} J_\ell(x_1) \\ \vdots \\ J_\ell(x_N) \end{bmatrix} \in \mathbb{R}^{(NV) \times d}
\end{equation}
with aggregate Fisher information $\mathcal{I}_N(v) = \frac{1}{\sigma^2} J_\text{stack}^\top J_\text{stack} = \frac{1}{\sigma^2}\sum_{i=1}^N J_\ell(x_i)^\top J_\ell(x_i)$. The null space of $\mathcal{I}_N(v)$ is:
\begin{equation}
\ker(\mathcal{I}_N) = \ker(J_\text{stack}) = \bigcap_{i=1}^N \ker(J_\ell(x_i))
\end{equation}
Any direction $u \in \bigcap_i \ker(J_\ell(x_i))$ satisfies $J_\ell(x_i) u = 0$ for all $i$, meaning it contributes zero information regardless of $N$. If the intersection of null spaces is non-trivial — which empirically holds when all prompts probe the same effective output subspace — then $\ker(\mathcal{I}_N) \neq \{0\}$ for all $N$ and the Cramér-Rao bound remains infinite along those directions. Non-identifiability is therefore not a small-sample problem: it is an intrinsic property of the model geometry that no amount of behavioral data can resolve.
 
\section{Intuitive Explanations}
\label{app:geometry}
% ------------------------------------------------------------
 
\subsection{Geometric intuition for null-space ambiguity}
 
Consider a simplified setting where a 3D steering vector $v \in \mathbb{R}^3$ affects 2D outputs $o \in \mathbb{R}^2$ through a projection matrix $J \in \mathbb{R}^{2 \times 3}$. By the rank-nullity theorem, $\dim(\ker(J)) = 3 - 2 = 1$. Concretely, if $J = [I_2 \mid 0]$ projects onto the first two coordinates, then $\ker(J) = \operatorname{span}\{(0, 0, 1)^\top\}$. Any vector of the form $v' = v + \beta (0, 0, 1)^\top$ produces identical outputs since:
\begin{equation}
J(v + \beta e_3) = Jv + \beta J e_3 = Jv + \beta \cdot 0 = Jv
\end{equation}
The output depends only on the projection of $v$ onto $\operatorname{row}(J)$, while the component in $\ker(J)$ is entirely invisible. Figure~\ref{fig:null_space_visual} illustrates this geometry: the observable output space sees only the $(v_1, v_2)$ projection, while $v_3 \in \ker(J)$ leaves no trace in any output. In language models, $\ker(J_\ell)$ occupies 86--94\% of the ambient activation space, so almost any perturbation to a steering vector goes undetected by behavioral evaluation.
 
\begin{figure}[h]
\centering
\begin{tikzpicture}[>=Stealth, every node/.style={font=\small}]
    \coordinate (O) at (0,0);
    \fill[blue!10, opacity=0.3] (-1.5,-2) -- (2.5,-2) -- (2.5,0.3) -- (-1.5,0.3) -- cycle;
    \node[blue!60!black, font=\scriptsize] at (1.8, -1.7) {Observable output space};
    \draw[->, thick] (O) -- (-1.2,-1.6) node[below left] {$v_1$};
    \draw[->, thick] (O) -- (2.2,0) node[right] {$v_2$};
    \node[font=\scriptsize, blue!60!black] at (0.5, -0.8) {$\text{row}(J)$};
    \draw[->, dashed, gray] (O) -- (0,1.8) node[midway, right, font=\scriptsize, gray] {$v_{\parallel}$};
    \draw[->, thick, red!70!black] (0,1.8) -- (0,3.2) node[above, red!70!black] {$v_3 \in \ker(J)$};
    \node[right, font=\scriptsize, red!70!black] at (0.1, 2.5) {(invisible)};
    \node[font=\footnotesize] at (0, -2.5) {Output plane sees only $(v_1, v_2)$ projection};
\end{tikzpicture}
\caption{Geometric intuition for null-space ambiguity. The output depends only on the $(v_1, v_2)$ projection (row space of $J$), while $v_3 \in \ker(J)$ is invisible to all outputs.}
\label{fig:null_space_visual}
\end{figure}
 
\subsection{Why more data does not resolve the ambiguity}
 
A natural conjecture is that collecting diverse prompts could reduce the null space by probing different output dimensions. Each observation provides $o_i = J_\ell(x_i) v + \eta_i$, and stacking $N$ observations gives a stacked Jacobian:
\begin{equation}
J_\text{stack} = \begin{bmatrix} J_\ell(x_1) \\ \vdots \\ J_\ell(x_N) \end{bmatrix} \in \mathbb{R}^{(NV) \times d}
\end{equation}
with null space $\ker(J_\text{stack}) = \bigcap_{i=1}^N \ker(J_\ell(x_i))$. In principle, diverse prompts that probe different output dimensions could reduce this intersection. However, directions $v_0$ satisfying $J_\ell(x) v_0 = 0$ for all $x$ are determined by model weights, not the prompt distribution. When all prompts probe the same effective output subspace, their Jacobians share a common null space and the intersection does not shrink regardless of how many prompts are added. Figure~\ref{fig:prompt_diversity_null_space} illustrates this: each prompt measures the same active dimensions, leaving the null-space intersection unchanged. The Fisher information along these null-space directions is identically zero, so by the Cramér-Rao bound the estimation variance is unbounded. Non-identifiability is not a small-sample artifact that more data can resolve — it is an intrinsic property of the model geometry.
 
\begin{figure}[h]
\centering
\resizebox{\columnwidth}{!}{
\begin{tikzpicture}[
node distance=6pt,
box/.style={draw, rounded corners, fill=blue!10, minimum width=0.9\columnwidth, minimum height=0.9cm, align=center, font=\ttfamily\footnotesize},
conclusion/.style={draw, rounded corners, fill=red!20, minimum width=0.9\columnwidth, minimum height=0.9cm, align=center, font=\ttfamily\footnotesize}
]
\node[box] (p1) {Prompt 1: measures dims [1,\ldots,139] $\rightarrow$ null space = [140,\ldots,2048]};
\node[box, below=of p1] (p2) {Prompt 2: measures dims [1,\ldots,139] $\rightarrow$ null space = [140,\ldots,2048]};
\node[below=of p2] (dots) {$\vdots$};
\node[box, below=of dots] (pn) {Prompt N: measures dims [1,\ldots,139] $\rightarrow$ null space = [140,\ldots,2048]};
\node[conclusion, below=of pn] (final) {Intersection of null spaces = [140,\ldots,2048] \quad (unchanged)};
\end{tikzpicture}
}
\caption{Why prompt diversity does not resolve null-space ambiguity. Each prompt probes the same effective subspace, leaving the null-space intersection unchanged. Numbers reflect Qwen2.5-3B ($d=2048$, rank 139).}
\label{fig:prompt_diversity_null_space}
\end{figure}
 
\subsection{Why orthogonal perturbations preserve semantic coherence}
 
Any perturbation $v_\perp \perp v$ decomposes as $v_\perp = v_{\perp, \text{row}} + v_{\perp, \text{null}}$, where $v_{\perp, \text{row}} \in \operatorname{row}(J_\ell)$ is observable through outputs and $v_{\perp, \text{null}} \in \ker(J_\ell)$ is invisible. For a random $v_\perp$, the expected fraction landing in the null space is approximately $\dim(\ker(J_\ell)) / d \approx 0.86$--$0.94$, meaning the bulk of any random perturbation is automatically undetectable. The remaining row-space component contributes an output change $J_\ell v_{\perp, \text{row}}$, but this is a random direction in the row space rather than a structured semantic shift. The output change from the original steering vector, $J_\ell v$, produces aligned and directional changes across the vocabulary that consistently shift the target trait. The row-space component of the perturbation, by contrast, produces diffuse and incoherent changes that average out across tokens without pushing any semantic dimension in a consistent direction.
 
To make the distinction concrete: if $J_\ell v$ is a strong wind blowing consistently northward, then $J_\ell v_{\perp, \text{row}}$ is turbulence. It may carry comparable energy but lacks directional coherence so the overall trajectory remains northward. This is why orthogonal perturbations leave semantic classifier scores nearly unchanged even when their norm is comparable to the steering vector itself.

\section{Prompt Construction Details}
\label{app:prompts}

We construct contrastive prompt pairs for each semantic trait using template-based generation. Table~\ref{tab:prompts} provides concrete examples for all five traits used in the empirical validation.  Held-out evaluation prompts are drawn from a separate set of 50 prompts
not used in vector extraction, covering in-distribution, topic-shift, and
genre-shift conditions for the multi-environment experiment.

\begin{table}[h]
\centering
\caption{Example contrastive prompt pairs per trait.}
\label{tab:prompts}
\small
\setlength{\tabcolsep}{4pt}
\renewcommand{\arraystretch}{1.2} % adds vertical spacing

\begin{tabularx}{\columnwidth}{lXX}
\hline
\textbf{Trait} & \textbf{Positive ($x^+$)} & \textbf{Negative ($x^-$)} \\
\hline
Formality
  & Draft a professional email summarizing quarterly performance.
  & Text a friend about how the quarter went. \\
Politeness
  & Respond diplomatically to criticism from a colleague.
  & Reply bluntly to someone who criticized you. \\
Sentiment
  & Write an enthusiastic review of a product you loved.
  & Write a negative review of a product you disliked. \\
Truthfulness
  & Provide an accurate factual summary of this event.
  & Make up plausible-sounding details about this event. \\
Agreeableness
  & Respond warmly and supportively to a colleague's proposal.
  & Push back sharply on a colleague's proposal. \\
\hline
\end{tabularx}
\end{table}

\begin{figure}[h]
\centering
\includegraphics[width=0.9\textwidth, trim=12 12 12 12, clip]{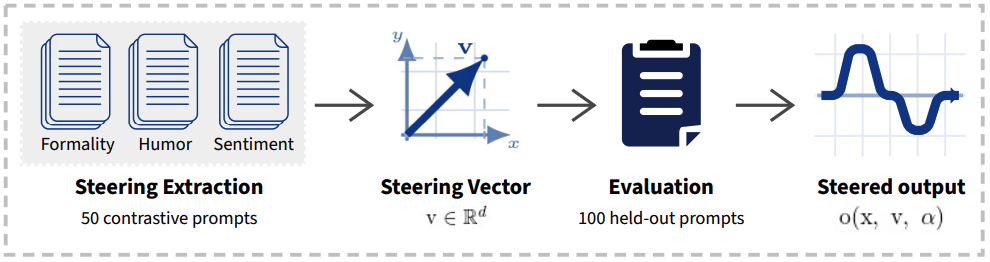}
\caption{Steering extraction and evaluation pipeline.}
\label{fig:test-prompts}
\end{figure}

Figure~\ref{fig:test-prompts} illustrates the pipeline used to extract and evaluate steering vectors. Contrastive prompt pairs are first used to compute a steering vector specific to a semantic trait. This vector is then applied to a separate set of held-out evaluation prompts to generate steered outputs.

\end{document}